\pdfoutput=1

\documentclass[11pt]{article}

\usepackage{acl} 

\usepackage{times}
\usepackage{latexsym}

\usepackage[T1]{fontenc}

\usepackage[utf8]{inputenc}

\usepackage{microtype}

\usepackage{inconsolata}

\usepackage{graphicx}

\usepackage{multirow}
\usepackage[inline]{enumitem}
\usepackage{graphicx}
\usepackage{gb4e}\noautomath
\usepackage{caption}
\usepackage{subcaption}
\usepackage{xcolor}

\title{EASSE-DE: Easier Automatic Sentence Simplification Evaluation \\ for German}

\author{
    Regina Stodden \\
  Department of Computational Linguistics \\ 
  Faculty of Arts and Humanities \\Heinrich Heine University Düsseldorf, Germany\\
  \texttt{regina.stodden@hhu.de} \\}

\begin{document}
\maketitle
\begin{abstract}
In this work, we propose EASSE-multi, a framework for easier automatic sentence evaluation for languages other than English. Compared to the original EASSE framework, EASSE-multi does not focus only on English. It contains tokenizers and versions of text simplification evaluation metrics which are suitable for multiple languages. In this paper, we exemplify the usage of EASSE-multi for German TS resulting in EASSE-DE.
Further, we compare text simplification results when evaluating with different language or tokenization settings of the metrics. Based on this, we formulate recommendations on how to make the evaluation of (German) TS models more transparent and better comparable. The code of EASSE-multi and its German specialisation (EASSE-DE) can be found in \url{https://github.com/rstodden/easse-de}.
\end{abstract}

\section{Introduction}
Automatic text simplification (TS) is a natural language processing (NLP) task that involves the development of algorithms and models to automatically transform complex textual content into more straightforward and accessible language. Manual or automatic evaluation is required to measure the quality of the generated simplifications. A good simplification should be grammatically correct, more simple and better readable than the original text and preserve the original meaning of it. For manual evaluation, people are asked to rate the extent of these three aspects for the generated simplification with respect to the original sentence. Because manual evaluation is very time-consuming, automatic metrics are used for a first quality check of sentence simplification models~\cite{alva-manchego-etal-2020-data}.
Compared to manual evaluation methods, automatic evaluation methods facilitate a quick assessment of the output of various text simplification models, making it feasible to compare and iterate on different approaches efficiently. Further, with the increasing mass of evaluation data of different model approaches, it becomes challenging to evaluate this large number of generated texts manually. Automatic evaluation methods allow researchers to scale up their assessments to handle large datasets effectively~\cite{alva-manchego-etal-2020-data}.

\citet{alva-manchego-etal-2019-easse} proposed an evaluation framework for easier automatic sentence simplification evaluation, called EASSE, to facilitate a comparison of TS models on existing test sets and on the same evaluation metrics as well as to unify the implementation of the evaluation metrics. EASSE is nowadays the common standard for evaluating English TS models. Although it is specified for only English TS evaluation, it is often also used to evaluate TS models of other languages, e.g., German (see, e.g., \citealt{trienes-etal-2022-patient}), Spanish (see, e.g., \citealt{gonzalez-dios-etal-2022-irekialfes}), French (see, e.g., \citealt{cardon-grabar-2020-french}), Spanish (see, e.g., ~\citealt{holmer-rennes-2023-constructing}) or on a multi-lingual benchmark~\cite{ryan-etal-2023-revisiting}. However, using EASSE on non-English texts raises some problems, e.g., the tokenizer is not adapted to the language of interest, the BERT-Score is evaluated on an English-only BERT model, and the readability scores are only designed for English.

In this paper, we present EASSE-multi, an adaptation of EASSE for languages other than English, to make the evaluation of non-English TS easier and more robust. We exemplify its usage for one language with several TS resources, i.e., German and the German EASSE variant, EASSE-DE. 
We further analyze the effects of different settings in EASSE-DE
on TS metrics when evaluating non-English texts (i.e., German texts). 

\section{Related Work}
\subsection{Automatic Evaluation}
In order to automatically evaluate text simplification, SARI~\cite{xu-etal-2016-optimizing} is the primary metric to measure the overall simplicity quality. In more detail, SARI compares a generated simplification sentence with the source sentence and several references to estimate the quality of the lexical simplification. Further, most often BLEU~\cite{papineni-etal-2002-bleu} and BERT-Score-Precision~\cite{bert-score} are utilized to measure the similarity or meaning preservation between the original text and the system-generated simplification. Following \citet{alva-manchego-etal-2021-un}, BERT-Score-Precision can also measure overall simplicity even if not implemented for this use case. Recently, the LENS score~\cite{maddela-etal-2023-lens} has been proposed to measure the overall simplification quality of English simplifications; it is a trainable score trained on human assessments and English complex-simple pairs.

Readability formulas such as FRE or FKGL~\cite{flesch-1948-new} are also often used to estimate the readability of the system output~\cite{alva-manchego-etal-2021-un}. For a syntactical simplification evaluation, SAMSA~\cite{sulem-etal-2018-semantic} has been proposed: SAMSA is a reference-less metric based on the annotation of semantic structures.

The reliability of these metrics for English TS evaluation has been questioned in research, e.g.,  see \citet{sulem-etal-2018-bleu}, \citet{tanprasert-kauchak-2021-flesch}, or \citet{alva-manchego-etal-2021-un}. Another issue with automatic metrics is that the reliability of the scores has only been evaluated against human annotations of English annotations and that the correlations are not yet reproduced or repeated in other languages. To facilitate the assessment of the reliability of the metrics for German texts, we gather available system outputs and automatic scores regarding them in EASSE. As long as SARI, BLEU, and BERT-Score are still common practices in TS research, we will use them in our analysis but are also open to replacing or extending the metrics.

Furthermore, \citet{stodden-kallmeyer-2020-multi} have shown that the way how English sentences are simplified differs from the German or Spanish ways. Hence, different simplification metrics might be required per language. An approach in this direction could be learnable metrics (per language) as LENS~\cite{maddela-etal-2023-lens}, BETS~\cite{zhao-etal-2023-towards} or MeaningBERT~\cite{beauchemin-etal-2023-meaningbert}, which are currently only applied to English texts.

\subsection{Original EASSE Package}
The original EASSE package~\cite{alva-manchego-etal-2019-easse} is designed to ease the automatic evaluation of English sentence simplification. It contains the implementation of automatic evaluation metrics, including SARI, BLEU, SAMSA, FKGL, and BERT-Score, as well as a linguistic feature analysis on the simplification pairs utilizing the TS-eval package by \citet{martin-etal-2018-reference}. EASSE also stores English TS test sets and outputs of English TS systems, as well as builds an evaluation report regarding all specified metrics of all specified TS models to facilitate the whole evaluation process. It is commonly used to evaluate TS system outputs in English and other languages.

\section{System Overview: EASSE-multi}
In order to make EASSE language-independent and more robust for evaluating texts of languages other than English, we are proposing EASSE-multi (and its German variant EASSE-DE in the next section).

Therefore, we add a language constant to EASSE-multi to specify the currently evaluated language (e.g., ``DE'' for German in EASSE-DE). We also add SpaCy to the list of possible tokenizers to allow tokenization specified for languages other than English  (see \autoref{sec-change-tok}).

The language constant also allows to choose language-specific evaluation metrics, e.g., readability metrics (see \autoref{sec-change-readability}), different models for BERT-Score (see \autoref{sec-change-metrics}) and multi-lingual linguistic feature extraction (see \autoref{sec-change-feature}).

\subsection{Tokenization}
\label{sec-change-tok}
The original EASSE version currently supports 13a tokenization or white-space split tokenization (presuming pre-tokenized data). To include the language component into tokenization, we added the tokenizers of SpaCy~\cite{spacy} and the extension Spacy-Stanza~\cite{qi-etal-2020-stanza}\footnote{\url{https://github.com/explosion/spacy-stanza}} as they currently support the tokenization of roughly 70 languages and also support linguistic annotations, e.g., part-of-speech tagging and dependency parsing, which will be relevant for the linguistic feature extraction.

\subsection{Metrics}
\label{sec-change-metrics}

Evaluation metrics for TS are mostly language-independent, e.g., SARI, or BLEU, as they are n-gram-based methods. However, the n-grams depend on tokenization, which differs from language to language (see previous section). Following \citet{bert-score}, BERT-Score can be used for a specific language (e.g., using the English-only model RoBERTa~\citep{liu-etal-2019-roberta}) or in a multi-lingual setting (e.g., using a multi-lingual model such as BERT-multilingual~\citep{devlin-etal-2019-bert}). 

In EASSE-multi, the usage of the metrics is optimized regarding the evaluated language, as based on the language constant, the tokenizer and the BERT-model are chosen to fit non-English languages better.

\subsection{Readability}
\label{sec-change-readability}
Readability scores and the LENS-Score~\cite{maddela-etal-2023-lens} are language-dependent, for the first due to included language-specific averages of word and sentence lengths and for the second due to training an evaluation score exclusively on English. 

As an extension of EASSE, we also added readability formulas for languages other than English to EASSE-multi, which have already been implemented in the textstat package. For example, common readability scores for German are the \citeauthor{amstad-1978-verstandlich}'s adaption on the Flesch Reading Ease (FRE) or the Vienna non-fictional text formulas~\cite{bamberger-vanecek-1984-lesen}. LENS has not been reproduced for other languages; hence, it makes no sense to include it in EASSE-multi.

Following the criticism of \citet{tanprasert-kauchak-2021-flesch} regarding readability metrics for TS evaluation, we follow their recommendation and include average sentence length, number of syllables and number of splits in our report. Hence, we add these features to the default report.

\subsection{Multi-lingual Feature Extraction}
\label{sec-change-feature}

As argued in \citet{tanprasert-kauchak-2021-flesch} and \citet{alva-manchego-etal-2019-easse}, we include a few linguistic features to get more insights into the system-generated simplification. For this, we are using the feature extraction toolkit of the reference-less quality estimation tool (further called TS-eval) by \citet{martin-etal-2018-reference} for the English analysis and its extended language-independent version TS-eval-multi by \citet{stodden-kallmeyer-2020-multi}. We decided to use TS-eval-multi for feature extraction and not the similar language-independent feature extraction toolkit called LFTK~\cite{lee-lee-2023-lftk} as both versions of TS-eval focus more on features for text simplification, whereas LFTK focuses more on features for readability assessment. The TS-eval package has also already been integrated into the evaluation package EASSE, which facilitates its extension to the multilingual TS-eval. Further, most of LFTK's implemented features only apply to English. In future work, TS-eval-multi could be extended with features of LFTK.  TS-eval-multi contains, for example, the parse tree height, cosine similarity between source and output based on pre-trained word embeddings, and length of phrases and clauses.

\subsection{Additional Resources}
The original EASSE framework also includes resources of English TS, i.e., English TS test sets, word lists, and system outputs of English TS models. With EASSE-multi, this component can be extended to the language of interest. We exemplify this with EASSE-DE and add only German resources (see \autoref{sec-exp}). However, the German resources can be easily replaced with resources of other languages.

\subsection{Recommended Setting}
At the moment, we cannot provide recommended settings per language except specifying the language constant, using SpaCy for tokenization, and using the multi-lingual BERT-Score. Further recommendations, for example, if case sensitivity is useful for the language of interest or determining which BERT version is more suitable for the language of interest, require more analysis which is out of the scope of this work. 
However, we recommend always naming which kind of settings have been used during evaluation as it can greatly influence the TS metrics. The settings should be reported in detail to ensure that the effect on the metric is due to the TS system and not the evaluation metrics' settings. 

Furthermore, it could be helpful to report the results of the baselines, e.g., src2src (source-to-source or complex-to-complex) or tgt2tgt (target-to-target or simple-to-simple). If the system outputs cannot be made available, it could help to verify on the gold data whether the applied evaluation method (e.g., in a replication experiment) is the same as the evaluation method used for an original experiment, as the results should be identical. Additionally, it could be helpful to re-evaluate the data comparing to. Therefore, we recommend making the system outputs publicly available (if the data is not restricted by license or copyright), e.g., as part of the EASSE-DE resources.

\section{EASSE-DE: Using EASSE-multi for German TS Evaluation}
\label{sec-exp}
We will exemplify the usage of EASSE-multi for one language, i.e., German, resulting in EASSE-DE. Therefore, we add German resources to EASSE-DE (see \autoref{sec-resources}), i.e., German sentence simplification test sets (see \autoref{sec-de-test-sets})), and available outputs of German TS systems regarding these test sets (see \autoref{sec-de-models}). Further, we analyze whether and to what extent differences exist when evaluating German text simplification with the original evaluation framework EASSE or its adaptation EASSE-DE (see \autoref{sec-compare-easse}). 

\subsection{German TS Resources}
\label{sec-resources}
\subsubsection{German TS Test Sets}
\label{sec-de-test-sets}
For a better overview of available test sets for German sentence simplification, we have added gold data, i.e., manually simplified complex-simple sentence pairs, to EASSE-DE. In more detail, EASSE-DE refers to nine test sets, i.e., ABGB\cite{meister-2023-abgb}, APA-LHA-OR-A2~\cite{spring-etal-2021-exploring}, APA-LHA-OR-B1~\cite{spring-etal-2021-exploring}, BiSECT~\cite{kim-etal-2021-bisect}, DEplain-APA~\cite{stodden-etal-2023-deplain}, DEplain-web~\cite{stodden-etal-2023-deplain}, TextComplexityDE~\cite{naderi-etal-2019-subjective}, GEOlino~\cite{mallinson-etal-2020-zero}, and Simple-German-Corpus~\cite{toborek-etal-2023-new}. We refer to \autoref{tab:test-set-sent} for more meta data of the test sets.

\begin{table*}[htb]
\resizebox{\textwidth}{!}{
\begin{tabular}{llllll|lll|lll}
\textbf{} &  &  &  &  &  & \multicolumn{3}{c|}{\textbf{complex}} & \multicolumn{3}{c}{\textbf{simple}} \\
\multicolumn{1}{l|}{\textbf{name}} & \textbf{target group} & \textbf{domain} & \textbf{size} & \textbf{\# ref.} & \textbf{n:m} & \textbf{FRE$\downarrow$} & \textbf{sent. len.$\uparrow$} & \textbf{word len.$\uparrow$} & \textbf{FRE$\uparrow$} & \textbf{sent. len.$\downarrow$} & \textbf{word len.$\downarrow$} \\ \hline
\multicolumn{1}{l|}{\textbf{ABGB}} & non-experts & law & 448 & 2 & 40\% & 42.75 & 24.85 & 1.83 & 44.6 & 22.39 & 1.89 \\
\multicolumn{1}{l|}{\textbf{APA\_LHA-or-a2}} & Non-native speaker & news & 500 & 1 & 6 \% & 44.7 & 20.2 & 1.92 & 69.55 & 11.27 & 1.78 \\
\multicolumn{1}{l|}{\textbf{APA\_LHA-or-b1}} & Non-native speaker & news & 500 & 1 & 8 \% & 43.7 & 20.48 & 1.93 & 62.6 & 12.82 & 1.83 \\
\multicolumn{1}{l|}{\textbf{BiSECT}} & people w. reading problems & politics & 753 & 1 & 100 \% & \textbf{8.55} & \textbf{30.24} & 2.01 & 35.85 & 15.72 & 1.98 \\
\multicolumn{1}{l|}{\textbf{DEplain-APA}} & Non-native speaker & news & 1,231 & 1 & 27 \% & 58.75 & 11.92 & 1.86 & 65.8 & 10.55 & 1.79 \\
\multicolumn{1}{l|}{\textbf{DEplain-web}} & mixed & web/mixed & 1,846 & 1 & 57 \% & 62.95 & 19.13 & 1.64 & \textbf{77.9} & 10.76 & \textbf{1.57} \\
\multicolumn{1}{l|}{\textbf{GEOlino}} & children & encyclopedia & 663 & 1 & 40 \% & 61.5 & 13.31 & 1.7 & 66.0 & 9.94 & 1.66 \\
\multicolumn{1}{l|}{\textbf{simple-german-corpus}} & mixed & web/mixed & 391 & 1 & 73 \% & 41.15 & 13.96 & 2.0 & 65.4 & \textbf{9.31} & 1.83 \\
\multicolumn{1}{l|}{\textbf{TextComplexityDE}} & Non-native speaker & encyclopedia & 250 & 1 & 83 \% & 28.1 & 27.75 & \textbf{2.08} & 51.2 & 14.17 & 1.9
\end{tabular}
}
\caption{Overview Test Sets for German Sentence Simplification which are included in EASSE-DE. Including the target group, domain, size in sentence pairs, number of references, percentage of $n:m$ alignments, word length measured in syllables, and sentence length measured in words.}
\label{tab:test-set-sent}
\end{table*}

\subsubsection{German TS Models}
\label{sec-de-models}
For German TS, a few models are available or reproducible, e.g., ZEST, by \citet{mallinson-etal-2020-zero}, sockeye by \citet{spring-etal-2021-exploring}, custom-decoder-ats by \citet{anschutz-etal-2023-language}, the few-shot approaches on BLOOM by \citet{ryan-etal-2023-revisiting}, or the mBART models by \citet{stodden-etal-2023-deplain}. A more detailed description and analysis of German TS models, including their reproduction, has been recently proposed by \citet{stodden-2024-reproduction}. The system outputs of all reproduced German TS models (see \autoref{tab:overview-models}) have been added to EASSE-DE \autoref{tab:overview-models} to facilitate a better comparison between existing models and models which will be newly proposed in future. 

\begin{table*}[htb]
\resizebox{\textwidth}{!}{
\begin{tabular}{p{4.5cm}|llllp{7cm}}
\textbf{System Name} & \textbf{Reference} & \textbf{Type} & \textbf{Training Data} & \textbf{\# Simp. Pairs} & \textbf{URL} \\ \hline\hline
\textbf{hda-etr} & \citet{siegel-etal-2019-aspects} & rule-based & - & - & \url{https://github.com/hdaSprachtechnologie/easy-to-understand\_language} \\\hline
\textbf{sockeye-APA-LHA} & \begin{tabular}[c]{@{}l@{}}\citet{spring-etal-2021-exploring} \& \\ \citet{ebling-etal-2022-automatic}\end{tabular} & seq2seq & \begin{tabular}[c]{@{}l@{}}APA-LHA OR-A2 \& \\ APA-LHA OR-B1\end{tabular} & \begin{tabular}[c]{@{}l@{}}8,455 \& \\ 9,268\end{tabular} & \url{https://github.com/ZurichNLP/RANLP2021-German-ATS} \\
\textbf{sockeye-DEplain-APA} & - & seq2seq & DEplain-APA & 10,660 & \url{https://huggingface.co/DEplain} \\
\textbf{mBART-DEplain-APA} & \citet{stodden-etal-2023-deplain} & \begin{tabular}[c]{@{}l@{}}fine-tuned\\ seq2seq\end{tabular} & DEplain-APA & 10,660 & \url{https://huggingface.co/DEplain/trimmed\_mbart\_sents\_apa} \\
\textbf{mBART-DEplain-APA+web} & \citet{stodden-etal-2023-deplain} & \begin{tabular}[c]{@{}l@{}}fine-tuned\\ seq2seq\end{tabular} & DEplain-APA+web & \begin{tabular}[c]{@{}l@{}}10,660 +\\  1,594\end{tabular} & \url{https://huggingface.co/DEplain/trimmed\_mbart\_sents\_apa\_web} \\
{\textbf{mT5-DEplain-APA}} & - & \begin{tabular}[c]{@{}l@{}}fine-tuned\\ seq2seq\end{tabular} & DEplain-APA & 10,660 & \url{https://huggingface.co/DEplain} \\
{\textbf{mT5-SGC}} & - & \begin{tabular}[c]{@{}l@{}}fine-tuned\\ seq2seq\end{tabular} & SGC & 4,430 & \url{https://huggingface.co/DEplain} \\\hline
\textbf{BLOOM-zero} & \citet{ryan-etal-2023-revisiting} & \begin{tabular}[c]{@{}l@{}}zero-shot\\ AR model\end{tabular} & - & - & \url{https://github.com/XenonMolecule/MultiSim} \\
\textbf{BLOOM-sim-10} & \citet{ryan-etal-2023-revisiting} & \begin{tabular}[c]{@{}l@{}}few-shot \\ AR model\end{tabular} & \begin{tabular}[c]{@{}l@{}}TCDE19 \& \\ GEOlino\end{tabular} & 200 \& 959 & \url{https://github.com/XenonMolecule/MultiSim} \\
\textbf{BLOOM-random 10} & \citet{ryan-etal-2023-revisiting} & \begin{tabular}[c]{@{}l@{}}few-shot\\ AR model\end{tabular} & \begin{tabular}[c]{@{}l@{}}TCDE19 \&\\  GEOlino\end{tabular} & 200 \& 959 & \url{https://github.com/XenonMolecule/MultiSim} \\\hline
\textbf{custom-decoder-ats} & \citet{anschutz-etal-2023-language} & \begin{tabular}[c]{@{}l@{}}AR model + \\ fine-tuned \\ seq2seq\end{tabular} & \begin{tabular}[c]{@{}l@{}}Simplified, monolingual \\German data \&\\  20Minuten\end{tabular} & \begin{tabular}[c]{@{}l@{}}544,467 \& \\ 17,905\end{tabular} & \url{https://huggingface.co/josh-oo/custom-decoder-ats}
\end{tabular}}
\caption{Overview of German TS models including training details (i.e., training data and size of training samples). Each line separates different model types. Adaptation from \citet{stodden-2024-reproduction}.}
\label{tab:overview-models}
\end{table*}

\subsection{Comparison of EASSE and EASSE-DE}
\label{sec-compare-easse}
In the following section, we present and analyse the metric scores when using EASSE or EASSE-DE, including different settings on three test sets of one TS model. 

\subsubsection{Method}
\paragraph{Evaluation Settings.}
In the comparative analysis, we focus on the settings in EASSE regarding i) language specification (i.e., English vs. German), ii) tokenization method (i.e., none vs 13a vs SpaCy), iii) BERT model version (i.e., RoBERTa-large vs BERT-base-multilingual-cased), iv) FRE version (English vs German). Due to their n-gram-based approach, we expect the tokenization method to have an effect on SARI and BLEU but not on BERT-Score-Precision.

\paragraph{German TS Test Sets.}
In the analysis, we evaluate on three available German TS test sets: DEplain-APA~\citep{stodden-etal-2023-deplain}, DEplain-web~\citep{stodden-etal-2023-deplain}, and TextComplexityDE~\citep{naderi-etal-2019-subjective}. These test sets are all manually simplified and manually aligned, and, therefore, we expect a higher simplification quality for them as for other test sets, e.g.,  BiSECT~\cite{kim-etal-2021-bisect}\footnote{BiSECT is generated using machine translation of English texts. Due to this augmentation strategy, the German version includes encoding errors.} or APA-LHA~\cite{spring-etal-2021-exploring}\footnote{The training and validation sets of APA-LHA are automatically aligned, and, hence, more faulty compared to manually aligned corpora.}. Further, these three test sets include texts of different domains (news, web, and Wikipedia), and their simplification addresses different target groups (non-native speakers and people with cognitive disabilities). Hence, they represent different kinds of simplifications and therefore seem to be a good choice for our analysis. 

\paragraph{German TS Model.}
Further, we have selected the generated simplifications of one model, i.e., mBART-DEplain-APA+web. Reasons for the choice of this model are that it is ready-to-use without additional examples, and, following \citet{stodden-2024-reproduction}, this model achieves the best BERT-Scores across several test sets. In comparison, the BLOOM models by \citet{ryan-etal-2023-revisiting} are few-shot models that require additional complex-simple pairs to generate simplifications.

\subsubsection{Results}
The results of the mBART-APA+web model with different settings are presented in \autoref{table-tok-results}.\footnote{To ensure that the effects are not due to the system but to the evaluation changes, we also add the results of the identity baseline (see \autoref{table-tok-results-identity}).}
In the following, we analyse the differences regarding tokenization, readability scores, multi-lingual BERT-Score, and multi-lingual feature extraction.

\begin{table}[htb]
\centering
\resizebox{0.95\columnwidth}{!}{
\begin{tabular}{l|ll|llll}
\textbf{\textbf{}} & \textbf{Tok.} & \textbf{\textbf{Lang.}} & \textbf{\textbf{BLEU$\uparrow$}} & \textbf{\textbf{SARI$\uparrow$}} & \textbf{BS-P$\uparrow$} & \textbf{FRE$\uparrow$} \\ \hline
\multirow{4}{*}{\begin{tabular}[c]{@{}l@{}}TCDE19 \\ (n = 250)\end{tabular}} & spacy & EN & \textbf{18.56} & \textbf{37.69} & 0.39 & \textbf{57.37} \\
 & spacy & DE & 17.75 & 37.37 & \textbf{0.55} & 43.65 \\
 & 13a & DE & 18.04 & 37.41 & \textbf{0.55} & 43.55 \\
 & none & DE & 16.04 & 37.47 & \textbf{0.55} & 43.65 \\ \hline
\multirow{4}{*}{\begin{tabular}[c]{@{}l@{}}DEplain-APA \\ (n = 1231)\end{tabular}} & spacy & EN & \textbf{30.59} & \textbf{34.79} & 0.48 & \textbf{78.25} \\
 & spacy & DE & 28.03 & 33.81 & \textbf{0.64} & 65.2 \\
 & 13a & DE & 28.37 & 33.92 & \textbf{0.64} & 65.2 \\
 & none & DE & 24.69 & 32.88 & \textbf{0.64} & 65.2 \\ \hline
\multirow{4}{*}{\begin{tabular}[c]{@{}l@{}}DEplain-web \\ (n = 1846)\end{tabular}} & spacy & EN & 18.37 & \textbf{34.21} & 0.27 & \textbf{76.52} \\
 & spacy & DE & 17.99 & 34.07 & \textbf{0.44} & 69.05 \\
 & 13a & DE & 18.17 & 34.10 & \textbf{0.44} & 69.05 \\
 & none & DE & 15.97 & 33.67 & \textbf{0.44} & 69.05
\end{tabular}
}
\caption{Scores of trimmed-mbart-DEplain-APA+web when using different language settings and tokenizers. }
\label{table-tok-results}
\end{table}

\paragraph{Tokenization.}
As expected, different tokenization methods (including language specification) affect the calculation of metrics used for TS evaluation. The last three rows in each block of \autoref{table-tok-results} show the differences in the scores when using different tokenization strategies. We can see that the BERT-score is always the same for all settings due to the sub-word tokenization in BERT. 

The FRE scores are also robust across all test sets when looking at the trimmed-mBART results, but in \autoref{results-identity} \autoref{table-tok-results-identity}, we see slightly more differences. The SARI scores also change slightly, i.e., to less than 1 point in all settings, whereas the differences in the BLEU scores range between 2 to 3 points in all test sets. In conclusion, when comparing one model against another with a slightly different evaluation setting (here, the tokenizer), even these small changes can be wrongly interpreted as an improvement of the model idiosyncrasy. However, it is only due to the different settings. Therefore, we recommend stating all settings chosen for evaluation for a more reliable comparison between models.

\paragraph{Readability Metrics.}
As can be seen in \autoref{table-tok-results}, the scores are quite different wrt. to FRE for the English and German settings (see first two rows in each block). The results are different due to the different constants of the formulas and their dependency on different tokenization and syllable splitting. When interpreting the readability scores, they also result in different categories: Following \cite{amstad-1978-verstandlich}, the simplifications with the English setting on DEplain-APA and DEplain-web can be described as ``ease'' whereas they are categorized as ``simple'' using the German setting.
In summary, the language adaptation of readability scores can make a noticeable difference when interpreting the simplification results. 

\paragraph{BERT-Score.}
As shown in the first two rows of each row-block in \autoref{table-tok-results}, changing the transformer model of the BERT-Score significantly affects the BERT-Score. The scores using the multi-lingual model are much higher than those using the only-English model. Hence, the choice of the BERT model seems to have a high effect on the TS evaluation.

\paragraph{Linguistic Feature Extraction.}
Due to space restrictions, we do not include analysis regarding linguistic feature extraction. We refer the interested reader to the system reports in our code repository\footnote{\url{https://github.com/rstodden/easse-de}}, which also include the feature extraction results.

\section{Discussion \& Conclusion }
We have proposed EASSE-multi, which facilitates easy evaluation of sentence simplification in multiple languages. Therefore, we have extended the original EASSE package with a language-constant tokenizer, language-dependent version of BERT-Score, and language-wise readability scores.

Further, we have exemplified using EASSE-multi for German TS evaluation in the form of EASSE-DE. In comparing the results generated by  EASSE and EASSE-DE, we have shown that it is important to consider the text's language when evaluating. Following that, we recommend using EASSE-DE over EASSE when evaluating German sentence simplification models as it includes language-sensitive evaluation metrics. Even if the scores per metric might be lower when using EASSE-DE than EASSE, we argue that these are more reliable due to the language-sensitive metrics. 

Further, we argue that it is unreliable to compare scores (maybe originating from different papers) as they might be generated by using different evaluation settings. Before making a comparison, we recommend verifying whether the same settings of the metric have been used in both experiments (the referenced and the new one). Otherwise, the differences in the scores might not be dependent on the model changes (which is the question of interest) but on, for example, different kinds of tokenization. 
Therefore, we strongly recommend always specifying the settings or, even better, the implementation of the metrics used for the evaluation, as it can have a huge impact on the reported scores.  We identified the following aspects which should be reported accompanied with automatic evaluation: 
\begin{enumerate*}
    \item language setting (e.g., EN, or DE) for features (e.g., BERT-Score, FRE, or word length),
    \item tokenizer (e.g., none, 13a, or SpaCy),
    \item lower casing (True or False),
    \item BERT-Score model (e.g., RoBERTA-large, mT5, or BERT-base-multilingual-cased)
\end{enumerate*}

\section{Future Work}

Even if most of the scores are language-independent or can be easily adapted to work for other languages, as shown previously, there still might be problems in using the same scores for different languages due to language idiosyncrasies and different simplification operations per language. 
Approaches in the direction of language-wise evaluation of non-English TS could be learnable metrics (per language) as already proposed for English, e.g., LENS, BETS, or MeaningBERT. In future work, we want to investigate learnable metrics for non-English languages to fit the language idiosyncrasies better and add them to EASSE-DE.

Further, we would like to extend EASSE-DE to include more German TS resources. We hope that EASSE-DE will be useful for German TS researchers and invite them to contribute their test sets or system outputs to EASSE-DE.

\newpage

\bibliographystyle{acl_natbib}
\bibliography{anthology,custom}

\newpage
\appendix
\section{Results of Identity Baseline}
\label{results-identity}
\begin{table}[htbp]
\centering
\resizebox{0.95\columnwidth}{!}{
\begin{tabular}{l|ll|llll}
\textbf{} & \textbf{\textbf{Tok.}} & \textbf{\textbf{Lang.}} & \textbf{\textbf{BLEU$\uparrow$}} & \textbf{SARI$\uparrow$} & \textbf{BS-P$\uparrow$} & \textbf{FRE$\uparrow$} \\ \hline
\multirow{4}{*}{\begin{tabular}[c]{@{}l@{}}TCDE19 \\ (n = 250)\end{tabular}} & spacy & EN & \textbf{28.22} & 15.31 & 0.37 & \textbf{39.16} \\
 & spacy & DE & 27.31 & 14.99 & \textbf{0.55} & 28.1 \\
 & 13a & DE & 27.49 & 15.05 & \textbf{0.55} & 28.0 \\
 & none & DE & 24.43 & 13.78 & \textbf{0.55} & 28.1 \\ \hline
\multirow{4}{*}{\begin{tabular}[c]{@{}l@{}}DEplain-APA \\ (n = 1231)\end{tabular}} & spacy & EN & \textbf{29.28} & \textbf{16.17} & 0.45 & \textbf{77.64} \\
 & spacy & DE & 26.89 & 15.25 & \textbf{0.63} & 58.75 \\
 & 13a & DE & 7.25 & 15.35 & \textbf{0.63} & 64.6 \\
 & none & DE & 23.33 & 13.75 & \textbf{0.63} & 58.75 \\ \hline
\multirow{4}{*}{\begin{tabular}[c]{@{}l@{}}DEplain-web \\ (n = 1846)\end{tabular}} & spacy & EN & 21.24 & 12.09 & 0.25 & \textbf{70.33} \\
 & spacy & DE & 20.85 & 11.93 & 0.42 & 62.95 \\
 & 13a & DE & 20.89 & 11.94 & 0.42 & 62.95 \\
 & none & DE & 18.82 & 10.9 & 0.42 & 62.95
\end{tabular}
}
\caption{Scores of identity baseline on three test sets when using different language settings and tokenizers.}
\label{table-tok-results-identity}
\end{table}

\hfill\newpage
\end{document}